\documentclass[sigconf, review=false]{acmart}

\usepackage{booktabs} 
\usepackage{multirow}
\usepackage{algorithm}
\usepackage{algorithmic}
\usepackage{amssymb}
\usepackage{subfigure}
\usepackage{balance}
\setcopyright{acmcopyright}

\usepackage{color}
\usepackage[normalem]{ulem}
\iftrue
\iffalse
\newcommand{\redliu}[1]{{\color{red}{[\textit{Hong: }\textbf{#1}]}}}
\newcommand{\dele}[1]{{\color{blue}{[\sout{#1}]}}}
\else
\newcommand{\redliu}[1]{#1}
\newcommand{\dele}[1]{\!}
\fi

\copyrightyear{2018}
\acmYear{2018}
\setcopyright{acmcopyright}
\acmConference[MM '18]{2018 ACM Multimedia Conference}{October 22-26, 2018}{Seoul, Republic of Korea}
\acmBooktitle{2018 ACM Multimedia Conference (MM '18), October 22-26, 2018, Seoul, Republic of Korea}
\acmPrice{15.00}
\acmDOI{10.1145/3240508.3240519}
\acmISBN{978-1-4503-5665-7/18/10}

\begin{document}
\title{Supervised Online Hashing via Hadamard Codebook Learning}
\author{Mingbao Lin$^{1,2}$, Rongrong Ji$^{1,2*}$, Hong Liu$^{1,2}$, Yongjian Wu$^{3}$}

\affiliation{\institution{$^1$Fujian Key Laboratory of Sensing and Computing for Smart City, Xiamen University, China} \institution{$^2$School of Information Science and Engineering, Xiamen University, China} \institution{$^3$Tencent Youtu Lab, Tencent Technology (Shanghai) Co., Ltd, China}}
\email{lmbxmu@stu.xmu.edu.cn, rrji@xmu.edu.cn, lynnliu.xmu@gmail.com, littlekenwu@tencent.com}

\begin{abstract}
In recent years, binary code learning, \textit{a.k.a.} hashing, has received extensive attention in large-scale multimedia retrieval.
It aims to encode high-dimensional data points into binary codes, hence the original high-dimensional metric space can be efficiently approximated via Hamming space.
However, most existing hashing methods adopted offline batch learning, which is not suitable to handle incremental datasets with streaming data or new instances.
In contrast, the robustness of the existing online hashing remains as an open problem, while the embedding of supervised/semantic information hardly boosts the performance of the online hashing, mainly due to the defect of unknown category numbers in supervised learning.
In this paper, we propose an online hashing scheme, termed \textbf{H}adamard \textbf{C}odebook based \textbf{O}nline \textbf{H}ashing (HCOH), which aims to solve the above problems towards robust and supervised online hashing.
In particular, we first assign an appropriate high-dimensional binary codes to each class label, which is generated randomly by Hadamard codes.
Subsequently, LSH is adopted to reduce the length of such Hadamard codes in accordance with the hash bits, which can adapt the predefined binary codes online, and theoretically guarantee the semantic similarity.
Finally, we consider the setting of stochastic data acquisition, which facilitates our method to efficiently learn the corresponding hashing functions via stochastic gradient descend (SGD) online.
Notably, the proposed HCOH can be embedded with supervised labels and is not limited to a predefined category number.
Extensive experiments on three widely-used benchmarks demonstrate the merits of the proposed scheme over the state-of-the-art methods.
The code is available at \url{https://github.com/lmbxmu/mycode/tree/master/2018ACMMM_HCOH}.
\end{abstract}

%
%

%
\ccsdesc[500]{Computing methodologies~Computer vision}

\keywords{Binary Code Learning; Large-scale Image Retrieval; Online Hashing; Hadamard Codesbook}

\maketitle

\textbf{\small{ACM Reference Format:}}\\
{\small{Mingbao Lin, Rongrong Ji, Hong Liu, Yongjian Wu. 2018. Supervised
Online Hashing via Hadamard Codebook Learning. In 2018 ACM Multimedia
Conference (MM '18), October 22-26, 2018, Seoul, Republic of Korea.
ACM, New York, NY, USA, 9 pages. https://doi.org/10.1145/3240508.3240519}}

\section{Introduction}
With the growth of data scales, hashing-based methods have attracted extensive research attentions in large-scale multimedia retrieval \cite{wang2018survey}, which merits in low storage and efficient computation on large-scale datasets.
In principle, most existing works aim to map high-dimensional data into a compact Hamming space, such that the original data similarity can be approximated via Hamming distance efficiently.
To this end, most effective hashing schemes are \redliu{``}data dependent\redliu{''}, which relies on modeling labeled or unlabeled data to learn discriminative binary codes \cite{weiss2009spectral, gong2013iterative,heo2012spherical,
liu2014discrete,liu2017ordinal,liu2012supervised,shen2015supervised,lu2017latent}.

However, such a setting is hardly workable for various real-world applications.
Many applications require the \redliu{search engine }\dele{retrieval system} to index streaming data online.
In contrast, most existing works in hashing adopt batch-based learning on the prepared training data, which is only suitable for fixed dataset.
While facing new data, \redliu{batch-based learning }\dele{batch-learning} has to accumulate all the available data and \redliu{re-learns}\dele{re-trains} all hash functions.
To handle this problem, advanced batch-based hashing \cite{liu2017ordinal,jiang2017asymmetric} has been proposed to perform multiple passes over the data.
\dele{However}\redliu{ Unfortunately}, the frequent data loading becomes a major performance bottleneck.
In order to address the above challenges, online hashing (OH) \cite{shalev2012online,wang2014online,crammer2004online} has been proposed to perform online learning of hash functions in an efficient way.
However, two open problems still exist\redliu{:}

Firstly, \redliu{most} OH methods require that the input data should be fed with pairs or batches \cite{leng2015online,huang2013online,fatih2017mihash}, while little works consider the case of a single datum input.
To tackle this problem, inspired by the Error Correcting Output Codes (ECOCs), Cakir \emph{et al.} proposed an online supervised hashing to solve such an extreme input \cite{cakir2017online}, which uses an SGD of the supervised hashing with error correcting codes.
\redliu{But}, the random construction for error correcting code will corrupt the model, which degenerates the retrieval performance.

Secondly, the unsupervised OH method \cite{leng2015online} can not make full use of the label information and suffers from low performance, which has to adopt a batch of training data to update the hash functions.
\dele{However}\redliu{ Meanwhile}, the performance of the existing supervised OH schemes \cite{huang2013online, cakir2015adaptive, cakir2017online} is still far from satisfying, most of which remain unchanged or even degenerate with the increase of streaming data, as reported in \cite{cakir2017online} and quantitatively demonstrated in Sec. \ref{sec4.4}.
Even though  the work in \cite{fatih2017mihash} solves this problem to a certain extend, its performance gain is accompanied by a time-consuming burden, due to the complex mutual information calculation between distance matrix and neighborhood indicator  (\emph{i.e.}, a O($T^2$) complexity where T is the number of batch size).
Therefore, an effective yet efficient OH scheme is in urgent need.

\begin{figure}[!tb]
\begin{center}
\includegraphics[height=0.35\linewidth]{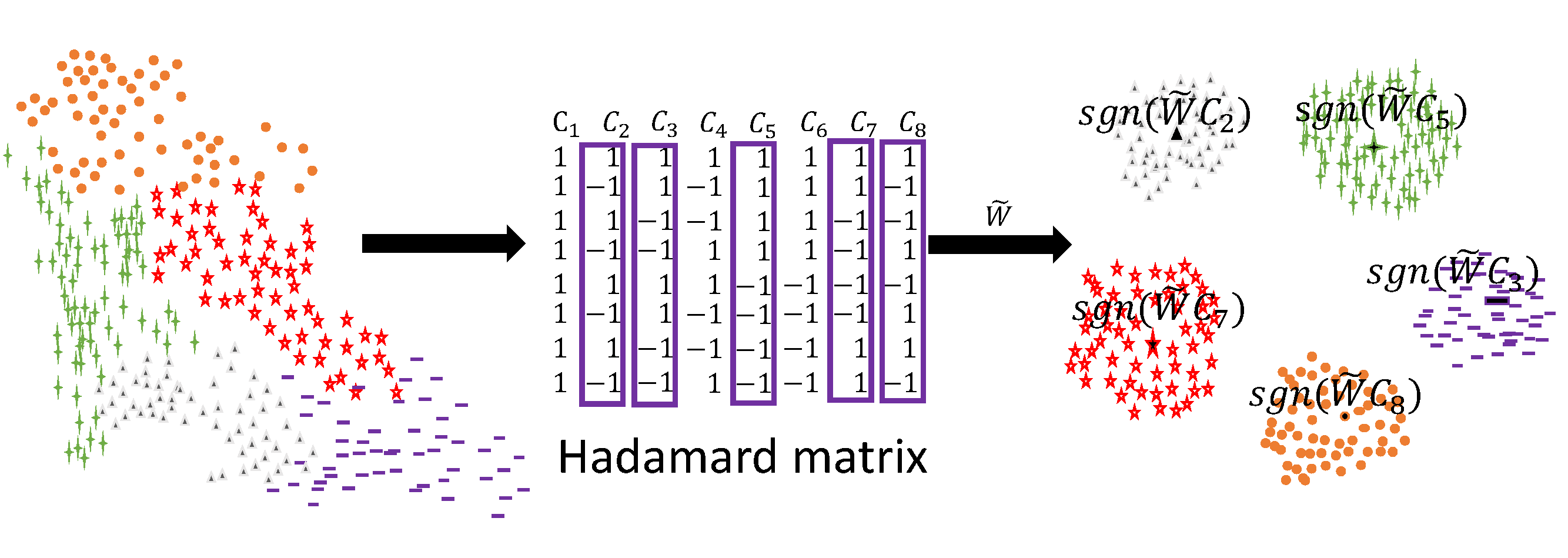}
\vspace{-1em}
\caption{ The framework of the proposed Hadamard Codebook based Online Hashing (HCOH). When new stream data from new class in left part is received, a column vector (codeword) in Hadamard matrix is sampled as a virtual \redliu{multi-label representation}\dele{label} to this class.
Otherwise, it shares the same virtual \redliu{multi-label representation}\dele{label} with instances falling in the same class (denoted with different shapes in different colors).
These virtual \redliu{representations}\dele{labels} form the codebook $C$ (purple boxes).
Then, a \redliu{random}\dele{randomized} Gaussian matrix $\tilde{W}$ is applied to reduce the  codeword  length to being consistent to the predefined hash bits.
In sum, HCOH explicitly aggregates the undiscriminating features in the original space into discerning Hamming space (in the right part).}
\label{framework}
\end{center}
\end{figure}

In this paper, we propose a simple and effective online hashing method, termed Hadamard Codebook based Online Hashing (HCOH), which aims to solve all aforementioned problems in a unified framework.
First, we generate the Hadamard matrix via it definition, from which a discrete codebook is randomly sampled.
Each codeword in such a codebook will be designated as the centroid of data sharing the same label, which can be used to conduct the learning of hash functions.
Second, we employ the locality sensitive hashing (LSH) \cite{gionis1999similarity} to reduce the  codeword  length to being consistent with the length of hash bits.
Therefore, HCOH can train the objective function by leveraging the difference between the codewords and the produced Hamming \redliu{codes, }\dele{space} and the hash functions are updated swiftly in an iterative manner with streaming data.
\redliu{Note that, both Hadamard matrix and the application of LSH can be efficiently applied online, because Hadamard matrix can be generated offline and the LSH is the data-independent encoding method with random projections. }

No extra time is spent during online learning by using this method, which differentiates our method from the existing OH method \cite{cakir2017online}, where the codebook has to be generated on-the-fly or to be in a fixed size.
Moreover, the proposed HCOH by nature enables the embedding of supervised labels by using discriminative Hadamard matrix.
Finally, in optimization, we show that the proposed HCOH needs only one instance to update each round, while most OH methods \cite{huang2013online,leng2015online,cakir2015adaptive,fatih2017mihash} need at least two instances.
Extensive experiments on three benchmarks, \emph{i.e.}, CIFAR-10, Places205, and MNIST, show that the proposed HCOH achieves better or competitive results to the state-of-the-art methods \cite{huang2013online,leng2015online,cakir2015adaptive,cakir2017online,fatih2017mihash}.

The main contributions of this work are as follows:
\begin{enumerate}
\item A codeword sampled from Hadamard matrix is introduced as the centroid of different class labels, which can be utilized to learn discriminative binary codes in online manner.
\item Each codeword can be used as the virtual multi-label representation, which serves as the supervised information to build our effective model for online learning.
\item The specific stochastic gradient descent (SGD) is derived to achieve efficient optimization for the proposed method.
\item The proposed scheme achieves competitive results compared with several state-of-the-art online hashing  \cite{huang2013online,leng2015online,cakir2015adaptive,cakir2017online,fatih2017mihash}.
\end{enumerate}
\dele{The rest of this paper is organized as follows: In Sec. \ref{sec2}, the related work is discussed. The details of the proposed HCOH and its optimization are presented in Sec. \ref{sec3}. Sec. \ref{sec4} reports our quantitative evaluations and analysis. Finally, we conclude this paper in Sec. \ref{sec5}.}

\section{Related Work} \label{sec2}
Recently, online hashing (OH) has received wide attention in online applications.
It merits in efficiently updating the hash functions by using the streaming data online, which can be further subdivided into two categories:
SGD-based OH methods \cite{huang2013online,cakir2015adaptive,fatih2017mihash,cakir2017online}, and matrix sketch-based OH methods \cite{leng2015online}.

For SGD-based online hashing, Online Kernel Hashing (OKH) \cite{huang2013online} is the first of its kind, which updates hash functions with an online passive-aggressive algorithm \cite{crammer2006online}.
OKH needs the new data to arrive in pairs with a similarity indicator, and the hash functions are updated via  gradient descent on the selected hashing parameters.
Similar to OKH, Adaptive Hashing (AdaptHash) \cite{cakir2015adaptive} uses a similar framework, in which data are fed with pairs and label similarity.
It defines a \dele{hinge-like loss}\redliu{ hinge loss} function and uses SGD to optimize model in online manner.
Furthermore, Cakir \textit{et al}. developed a more general two-step OH framework, \textit{a.k.a.} Online Supervised Hashing (OSH) \cite{cakir2017online}.
\redliu{In details of OSH, }\dele{First} Error Correcting Output Codes (ECOCs) \cite{dietterich1995solving} are \redliu{first} generated as the codebook, in which each codeword is further assigned to \redliu{each}\dele{per} class.
Then, an exponential loss is developed to replace the $0\backslash 1$ loss, which is optimized via SGD to ensure the learned hash functions to fit the binary ECOCs.
In \cite{fatih2017mihash}, Online Hashing with Mutual Information (MIHash) was proposed by giving an image  along with its neighbors and non-neighbors.
It targets at optimizing the mutual information to reduce the ambiguity in the induced neighborhood structure in the Hamming space.
Therefore, the whole framework can be optimized via SGD on mini-batch data.

For sketch-based online hashing, the main motivation comes from the idea of \redliu{``}data sketching\redliu{''}, which preserves the main property of a dataset with a significantly smaller size \cite{clarkson2009numerical, liberty2013simple}.
Leng \textit{et al.} proposed Online Sketching Hashing (SketchHash) \cite{leng2015online} where an online sketching with zero mean is developed to efficiently update hash codes online.
An efficient variant of SVD decomposition (denoted as RSVD) \cite{sproston1980matrix} is employed to obtain the hash functions.
Although SketchHash \cite{leng2015online} only requires $O(dl)$ space complexity to store and perform calculations on the streaming data, its time complexity is $O(ndl+dl^2)$ to yield hash functions, where $n$ is the data size, $d$ is the data dimension, and $l$ denotes the sketching size satisfying $l<d\ll n$.
\redliu{Moreover,} FasteR Online Sketching Hashing (FROSH) \cite{Chen2017FROSHFO} was developed to further reduce the training time to $O(nl^2+nd)$.
FROSH employs the independent Subsampled Randomized Hadamard Transform (SRHT) on different small data chunks to make the sketching compact and accurate, as well as to accelerate the sketching process.

\section{The PROPOSED METHOD} \label{sec3}
\subsection{Notations}
In this section, we introduce the proposed Hadamard Codebook based Online Hashing (HCOH) in details.
We first give notations used in the rest of this paper.
We define $X = [x_1, ..., x_n]$ as a set of $n$ training data with the corresponding labels $Y = \{y_1, ..., y_n\}$, where each $x_i \in \mathbb{R}^d$ is the $i$-th $d$-dimensional feature with label $y_i \in \mathbb{R}$.
Let $\mathcal{H}_1 = \{-1,1\}^{r}$ be the $r-$dimensional Hamming space.

The goal of hashing is to assign each instance a binary code in $\mathcal{H}_1$, such that similarities in the original space are preserved in the Hamming space.
This is achieved by learning a collection of hash functions $F = \{f_1, ..., f_k,...,f_{r}\}$, where each function $f_k : \mathbb{R}^d \rightarrow \{-1, +1\}$ is to generate one bit code.
As for online hashing, the hash functions $F$ are continuously updated from the input streaming data.
Following \cite{Chen2017FROSHFO}, we mainly consider the linear projection-based hash function, where hash function for each bit is defined as:
$$f_{k}(x)= \mathrm{sgn}(w_{k}^Tx+b_{k})=
\begin{cases}
$ 1$,& \mathrm{if}~w_{k}^Tx+b_{k} \geq 0, \\
$-1$,& \mathrm{otherwise},
\end{cases}$$
where $w_{k}\in \mathbb{R}^{d}$ is a parameter vector and $b_{k}\in \mathbb{R}$ is a bias term.
Consequently, the hash codes for $X$ can be presented as
\begin{eqnarray}
	F(X) = [f_1(X)\redliu{,} ...\redliu{,} f_{r}(X)] = \mathrm{sgn}(W^TX + b) \subset \mathbb{R}^{r \times n},
\end{eqnarray}
where $W = [w_1,w_2,...,w_{r}] \in \mathbb{R}^{d \times r}$ and $b=[b_1;b_2;...;b_{r}] \in \mathbb{R}^{r}$.

\subsection{Online Hashing Formulation}
To update the hash functions $F$ from streaming data online, the current mapping matrix $W^t$ and bias vector $b^t$ are learned on the $t$-th round input streaming data $X^t = [x^t_1, ..., x^t_{n^t}] \in \mathbb{R}^{d\times n^t}$ with their corresponding class labels $Y^t = \{y_1^t, ..., y^t_{n^t}\}\in \mathbb{R}^{n^t}$.
As mentioned, the existing online hashing methods \cite{leng2015online,huang2013online,fatih2017mihash} requires the input data to be paired or batched, \emph{i.e.}, $n^t \geq 2$, for dynamic updating.
In this paper, we \dele{first} break through such a limitation by updating the model using only one instance via stochastic gradient descend, which is experimentally demonstrated to be very effective.

To this end, we first revisit SGD-based online hashing, \textit{i.e.}, an SGD version of the supervised hashing with error correcting codes (ECC) \cite{cakir2017online}, known as \redliu{``}codebook\redliu{''}.
Each vector, known as \redliu{``}codeword\redliu{''} in this codebook is assigned to the data falling into the same label.
SGD-based hashing employs a \redliu{$0\!-\!1$} loss function, which outputs either $1$ or $0$ to indicate whether the binary code generated by the existing hash functions is matched to the codeword.
After replacing the \redliu{$0\!-\!1$} loss with a convex loss function and relaxing the  sign function, SGD is applied to minimize the loss and update the hash functions online.
However, such methods cannot guarantee a constant loss upper bound \cite{huang2013online}, which is caused by the random construction of the codebook for class label.
To compensate, a boosting scheme that considers previous mappings when updating each hash function is used to handle the error-correlation, which however needs more training time for each round input.

To solve the above \redliu{mentioned} problems, we argue that a better ECC \dele{codebook} is the key for robust and efficient online hashing.
The basic idea of ECC stems from the model of signal transmission in the communication field \cite{peterson1972error}.
\redliu{Recently, ECC has become one of the most widely used strategies for dealing with multi-class classification problems, which contains both encoding and decoding phases.
In the encoding phase, an encoding matrix (codebook) $C \in \{-1,1\}^{r^* \times N}$   decouples an $N$-class classification problem into $r^*$ binary-classification (bi-classification) problems }\cite{liu2016joint}.
\dele{It contains both encoding and decoding phases.
In the encoding phase, an encoding matrix (codebook) $C \in \{-1,1\}^{r^* \times N}$ decomposes an $N$-class classification problem into $r^*$ binary-classification (bi-classification) problems.}
That is, each column (codeword) of the matrix $C$ represents a class sample, each row represents a virtual category, and each original class can be approximated by a series of virtual categories \redliu{in the decoding phase}.
Therefore, \redliu{we argue that ECC can also help to solve the existing problems of online hashing, where} the hash functions can be seen as a set of bi-classification models, and each $f_k(x_i) = 1$ means a given $x_i$ belongs to the $k$-th virtual \dele{class}\redliu{catogory} and vice versa.

Following the above definitions, we consider the linear regression to \dele{evaluate}\redliu{build} each bi-classification model at the $t$-th round with the $i$-th new data point $x_i^t$:
\begin{equation}   \label{ori_sin_loss}
l(x_i^t;W^t) =  \|F(x_i^t) - c_{J(x_i^t)}\|_{\mathcal{F}}^2,
\end{equation}
where $c_i$ is the $i$-th column of matrix $C$, and $J(x_i^t)$ returns the class label of $x_i^t$, and $\|\cdot\|_{\mathcal{F}}$ is the Frobenius norm of the matrix.
Therefore, the overall objective function can be rewritten as:
\begin{align}\label{ori_total_loss}
		\min_{X^t;W^t,b^t}\redliu{\!\!\!\! L}(W^t,b^t) &\redliu{\!=\!}\frac{1}{n} \sum_{i=1}^n l(x_i^t;W^t,b^t) \redliu{\!=\!}\frac{1}{n} \sum_{i=1}^n\|F(x_i^t) - c_{J(x_i^t)}\|_{\mathcal{F}}^2 \nonumber \\
      &=\frac{1}{n} \|F(X^t)- C_{J(X^t)} \|_{\mathcal{F}}^2,
\end{align}
where \dele{function $J(x_i^t)$ returns the virtual label by giving data $x_i^t$, and} $C_{J(X^t)}\in \mathbb{R}^{r^* \times n^t}$ is the virtual \redliu{multi-}label representation matrix, of which the rows represent the virtual labels and the columns represent the samples.

However, the length of hash bit may not be the same with the length of codeword, \emph{i.e.}, $r \neq r^*$, which makes Eq. \ref{ori_total_loss} hard to be directly optimized.
To handle this problem, we further use the locality sensitive hashing (LSH) to transform the virtual labels to obtain the same length of binary codes to \redliu{learn} the hash functions.
As proven in \cite{ding2016defense}, LSH preserves the semantic similarity among label-based representation, since the collision probability between the binary codes of two labels is \redliu{less than} 0.5 if they are dissimilar, otherwise the collision is greater than 0.5.
As a result, we reformulate Eq. \ref{ori_total_loss} by LSH-based random hashing as:
\begin{equation}
	\min_{W^t,b^t} \frac{1}{n} \Big\|F(X^t)- \mathrm{sgn}\big( \tilde{W}^TC_{J(X^t)}\big) \Big\|_{\mathcal{F}}^2,\label{eq4}
\end{equation}
where $\tilde{W} \in \mathbf{R}^{r^* \times r}$ is the random Gaussian matrix.

\subsection{Hadamard Codebook}  \label{sec3.3}
As above, the key issue falls in the construction of encoding matrix \redliu{(codebook)} $C$.
Towards learning an optimal matrix $C$ via ECC method, the following principles should be satisfied in traditional ECC:
1) Maximize the Hamming distance between each row, which allows strong error-correction ability.
2) Maximize the Hamming distance between each column, which ensures significant difference between the classifiers.
However, the existing ECC construction schemes are too complex for generating long hash bits, which hinders their practical applications.

To solve this problem, we use the classical Hadamard codes in the communication system to construct ECC, which satisfies the above two principles  \cite{horadam2012hadamard}.
In particular the Hadamard is an $n$-order orthogonal matrix, \textit{i.e.}, both its row vectors and columns vectors are \redliu{pair-wise} orthogonal, which by nature satisfies principles $1)$ and $2)$, respectively.
And its elements are either $+1$ or $-1$. That is:
\begin{equation} \label{hadamard_definition}
CC^T = nI_n, ~\mathrm{or} ~ C^TC=nI_n,
\end{equation}
where $I_n$ is an $n-$order identity matrix.

Though the existence of Hadamard matrices of other orders \cite{paley1933orthogonal, williamson1944hadamard,goldberg1966hadamard,ockwig2005reticular}, we utilize $2^k$-order Hadamard matrices in this paper, which can achieve satisfactory performances as shown in Sec. \ref{sec4}.
To construct the $2^k$-order Hadamard matrices, the entry in the $i$-th row and the $j$-th column can be defined as:
\begin{equation}
\redliu{C}_{ij} = (-1)^{(i-1)\times(j-1)}.
\end{equation}

Based on the definition of Hadamard matrix in Eq. \ref{hadamard_definition}, vectors in a Hadamard matrix are linearly independent.
That is, Hadamard matrices can be well utilized as a discriminative set in Hamming space, which can further guide the learning of hash functions.

By the definition of $2^k$-order Hadamard matrix, we set the coding length $r^*$ as follows:
\begin{equation}   \label{r2}
\begin{aligned}
r^* = \min \{l|l=2^k, l\geq r, l \geq |Y|, k=1,2,3,...\},
\end{aligned}
\end{equation}
where $|Y|$ is the number of class labels in the dataset.
Therefore, we construct the square encoding matrix as $C_{r^*}\in \{-1,1\}^{r^* \times r^*}$.
If a new data with new label is received, we randomly and non-repeatedly select a column representation to construct a virtual \redliu{multi-}label vector for this data.
Otherwise, the virtual label previously assigned to the instances with the same label is given.
Such vectors are further aggregated to construct the encoding matrix $C$.
Therefore, our scheme does not need to predefine the category number of the dataset.
The detailed framework can be shown in Fig. \ref{framework}.

\subsection{Learning Formulation}
By giving the encoding matrix $C$ \redliu{as defined in Sec. \ref{sec3.3}}, we aim to optimize the objective function in Eq. \ref{eq4}.
However, the sign function $\mathrm{sgn}(\cdot)$ is non-smooth and non-convex, which makes the standard optimization method infeasible for the proposed model.
Following the work in \cite{liu2017ordinal}, we relax the hash function $F(X) = \mathrm{sgn}(W^TX + b)$ as follows:
\begin{equation}
\hat{F}(X) = \tanh(W^TX + b),
\end{equation}
where $\tanh(\cdot)$ is the hyperbolic tangent function that transforms the discrete values $\{-1, +1\}$ to continuous values $(-1, +1)$.

Based on this relaxation, Eq. \ref{ori_sin_loss} can be reformulated as:
\begin{equation}   \label{after_sin_loss}
\hat{\redliu{L}}(x_i^t;W^t, b^t) = \|\hat{F}(x_i^t) - c_{J(x_i^t)}\|_{\mathcal{F}}^2,
\end{equation}
Ideally, SGD optimization can be used to carry out in an iterative way for Eq. \ref{eq4}, where the partial derivative of $\hat{L}$ with regard to $W$ and $b$ can be derived as:
\begin{equation}
\begin{aligned}
W^{t+1} \leftarrow W^t -  \eta^t\frac{\partial \hat{L}(W^t)}{\partial W^t},
\end{aligned}
\end{equation}
\begin{equation}
\begin{aligned}
b^{t+1} \leftarrow W^t - \eta^t\frac{\partial \hat{L}(b^t)}{\partial b^t},
\end{aligned}
\end{equation}
where $\eta^t$ is a positive learning rate at the $t$-th round, and the derivative of $\hat{L}(W^t)$ with respect to $W^t$ is
\begin{equation}  \label{after_deriv_total}
\begin{aligned}
\frac{\partial \hat{L}(W^t)}{\partial W^t} = \frac{2}{n^t}X^t\Big(\big(\tanh({W^t}^TX^t+b^t) - \mathrm{LSH}(X^t)\big) \odot P\Big),
\end{aligned}
\end{equation}
and the derivative of $\hat{b^t}$ with respect to $b^t$ is
\begin{equation}  \label{derive_b}
\begin{aligned}
\frac{\partial \hat{L}(b^t)}{\partial b^t} = \frac{2}{n^t}\Big(\big(\tanh({W^t}^TX^t+b^t) - \mathrm{LSH}(X^t)\big) \odot P\Big),
\end{aligned}
\end{equation}
where \redliu{$P= \big(1- tanh({W^t}^TX^t+b^t)\big) \odot tanh({W^t}^TX^t+b^t)$}, $\odot$ denotes the element-wise product, and $\mathrm{LSH}(X^t) = \mathrm{sgn}\big( \tilde{W}^TC_{J(X^t)}\big)$.
%
\redliu{We summarize our proposed HOCH in Alg.\ref{alg1}.
}

\begin{algorithm}[t]
\caption{ Hadamard Codebook Hashing based Online Hashing (HCOH)}
\renewcommand{\algorithmicrequire}{\textbf{Input:}}
\renewcommand{\algorithmicensure}{\textbf{Output:}}
\begin{algorithmic}[1]
\REQUIRE
Training data set $D$ with feature space $X$ and label space $Y$, the number of hash bits $r$, the learning rate $\eta$, the total number of streaming data batches $L$.
\ENSURE
The hash codes $B$ for training space $X$ and the projection coefficient  matrix $\mathbf{W}$.\\
\STATE
Initialize $W_1$ and $b_1$ with the normal Gaussian distribution.
\STATE
Set the value of $r^*$ by Eq. \ref{r2}.
\STATE
Generate Hadamard matrix as stated in Sec. \ref{sec3.3}.

\IF{$ r = r^* $}
\STATE
Set $\tilde{W}$ as an identity matrix.
\ELSE
\STATE
Randomize $\tilde{W}$ from Gaussian distribution.
\ENDIF

\FOR {$t=1 \to L$}
    \STATE Update $W^t$ by Eq. \ref{after_deriv_total}.
    \STATE Update $b^t$ by Eq. \ref{derive_b}.
\ENDFOR
\STATE
Compute $B = sgn(W^TX + b).$
\end{algorithmic}
\label{alg1}
\end{algorithm}

\begin{table*}[]
\centering
\caption{\textit{m}AP and Precision@500 Comparisons on CIFAR-10 with 8, 16, 32, 64 and 128 bits. }
\label{cifar-map-precision500}
\begin{tabular}{c|ccccc|ccccc}
\hline
\multirow{2}{*}{Method} & \multicolumn{5}{c|}{\textit{m}AP}                   & \multicolumn{5}{c}{Precision@500}          \\
\cline{2-11}
                        & 8-bit & 16-bit & 32-bit & 64-it & 128-bit & 8-bit & 16-bit & 32-bit & 64-bit
                        &128-bit \\
\hline
OKH                     &0.100  &0.134   &0.223   &0.268  &0.350    &0.100  &0.179   &0.361   & 0.431
                        &0.510  \\
\hline
SketchHash              &0.248  &0.301   &0.302   &  -    &   -     &0.348  &0.433   &0.450   &   -
                        & -     \\
\hline
AdaptHash               &0.116  &0.138   &0.216   &0.305  &0.293    &0.129  &0.182   &0.357   &0.464
                        &0.467         \\
\hline
OSH                     &0.123  &0.126   &0.129   &0.127  &0.125    &0.138  &0.150   &0.150   &0.154
                        &0.157  \\
\hline
MIHash                  &0.512  &0.640   &0.675   &0.667  &0.664    &0.560  &0.703   &0.744   &0.739
                        &0.745    \\
\hline
\hline
HCOH                    &\textbf{0.536} &\textbf{0.698}   &\textbf{0.688}   &\textbf{0.724}  &\textbf{0.734}
                        &\textbf{0.636}  &\textbf{0.752}   &\textbf{0.756}   &\textbf{0.772} &\textbf{0.779}
\\
\hline
\end{tabular}
\end{table*}

\begin{figure*}[!t]
\begin{center}
\includegraphics[height=0.205\linewidth]{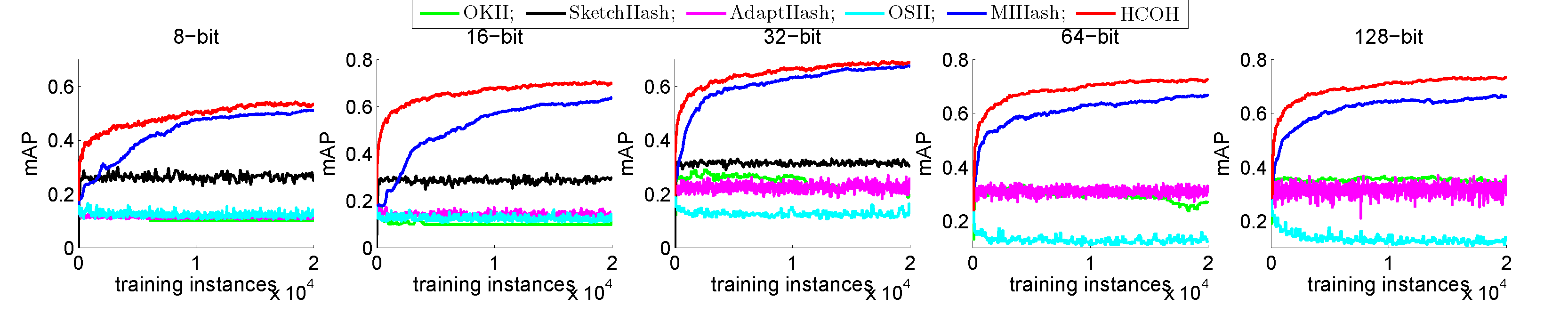}
\vspace{-2em}
\caption{\label{cifar-map} \textit{m}AP performance with respect to different number of training instances on CIFAR-10.}
\end{center}
\end{figure*}
\begin{figure}[!t]
\begin{center}
\includegraphics[height=0.63\linewidth]{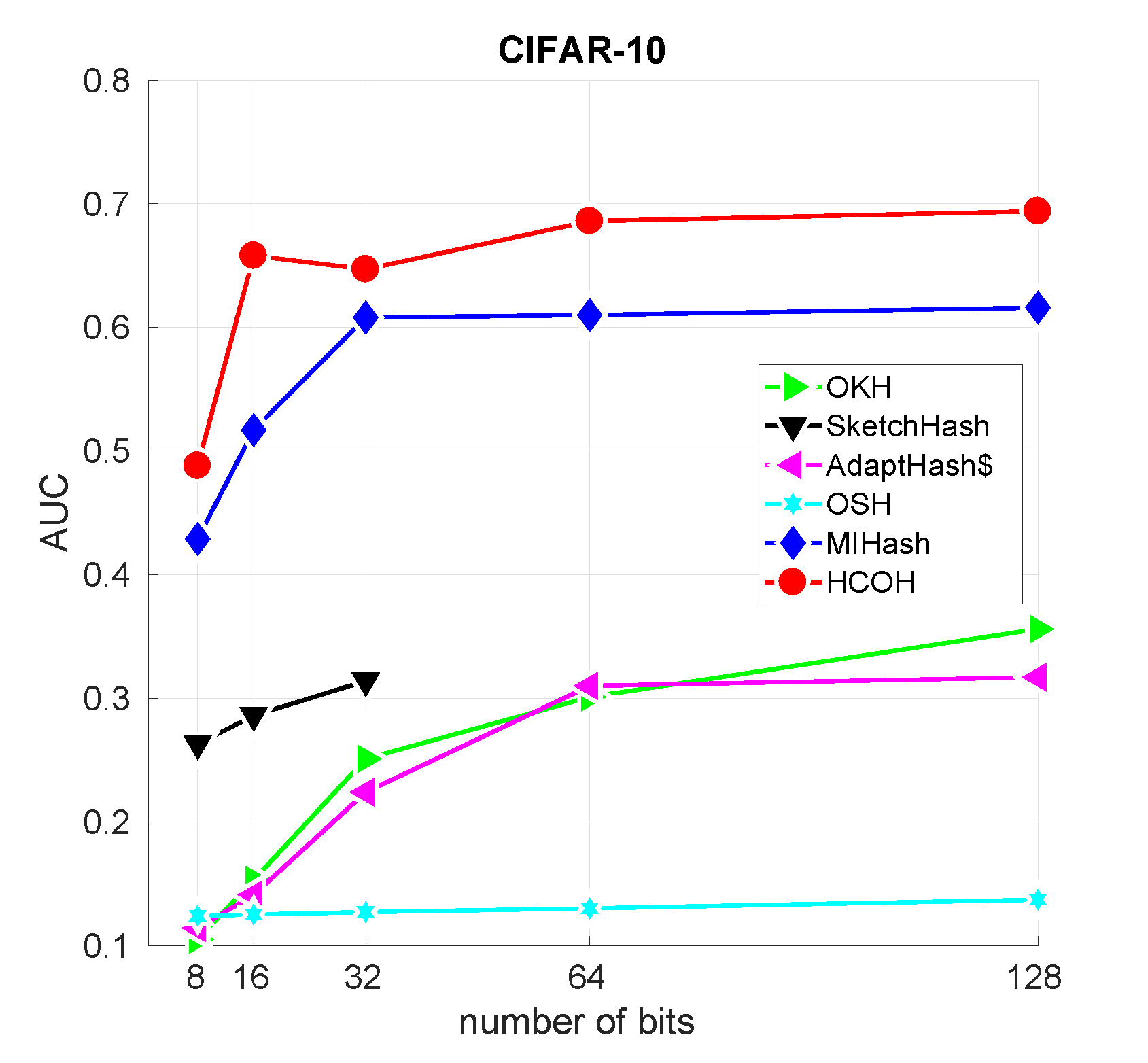}
\caption{\label{auc_cifar} AUC performance with respect to different lengths of hash bits on CIFAR-10.}
\end{center}
\end{figure}

\section{EXPERIMENTS} \label{sec4}
In this section, we report our quantitative experiments to verify the effectiveness and efficiency of the proposed method.
We run large-scale image retrieval experiments on three datasets, \textit{i.e.}, CIFAR-10 \cite{krizhevsky2009learning}, Places205 \cite{zhou2014learning}, and MNIST \cite{lecun1998gradient}.
\subsection{Datasets \label{datasets}}

\textbf{CIFAR-10} is a widely-used dataset for image classification and retrieval.
It contains 60K samples from 10 categories represented by 4096-dimentional CNN features extracted from the \textit{fc-7} layer of a VGG-16 network \cite{simonyan2014very} pre-trained on ImageNet.
As in \cite{fatih2017mihash}, the entire dataset is partitioned into two parts: a retrieval set with 59K samples, and a test set with 1K samples.
And a random subset of 20K images from the retrieval set is used for learning hash functions.

\textbf{Places205} is a 2.5-million image set where each instance belongs to one of 205 scene categories.
Following \cite{fatih2017mihash, cakir2017online}, features are pre-computed from the \textit{fc-7} layer of an AlexNet \cite{krizhevsky2012imagenet}, and then further reduced by PCA to a 128-dimension vector.
For each category, 20 instances are randomly sampled to form a test set, and the remaining are used to form a retrieval set.
We sample a subset from the retrieval set with 100K images at random for learning hash functions.

\textbf{MNIST} dataset contains 70K images of handwritten digits with 10 classes. Feature vectors are presented by $28 \times 28 = 784$ normalized original pixels.
We construct the test set by sampling $100$ instances from each class and the other are used as retrieval set.
From the retrieval set, $20K$ instances are adopted to learn the hash functions.

\subsection{Evaluation Protocols and Baselines}
As for the evaluation protocols, we adopt the widely-used mean Average Precision (\textit{m}AP) at varying bit lengths, as well as the mean precision of the top $500$ retrieved samples (denoted as Precision@500).
We also report the \textit{m}AP vs. numbers of training instances, as well as its corresponding area under the \textit{m}AP curve (denoted as AUC).
Due to the large scale of Places205 benchmark, it is very time-consuming to compute \textit{m}AP.
Following \cite{fatih2017mihash}, we only compute \textit{m}AP on the top $1000$ retrieved samples (Denoted as \textit{m}AP@1000).

We compare our method with five state-of-the-art online hashing algorithms, \textit{i.e.}, Online Kernel Hashing (OKH) \cite{huang2013online}, Online Sketching Hashing (SketchHash) \cite{leng2015online}, Adaptive Hashing (AdaptHash) \cite{cakir2015adaptive}, Online Supervised Hashing (OSH) \cite{cakir2017online} and Online Hashing with Mutual Information (MIHash) \cite{fatih2017mihash}.
Source codes of all these methods are available publicly.
Our model is implemented with MATLAB.
Training is done on a standard workstation with a \redliu{3.6GHz} Intel Core I7 4790 CPU and 16G RAM.
All the experimental results are averaged over three runs.

\begin{table*}[]
\centering
\caption{\textit{m}AP@1000 and Precision@500 Comparisons on Places205 with 8, 16, 32, 64 and 128 bits.}
\label{places-map-precision500}
\begin{tabular}{c|ccccc|ccccc}
\hline
\multirow{2}{*}{Method} & \multicolumn{5}{c|}{\textit{m}AP@1000}                   & \multicolumn{5}{c}{Precision@500}          \\
\cline{2-11}
                        & 8-bit & 16-bit & 32-bit &64-bit & 128-bit & 8-bit & 16-bit & 32-bit & 64-bit
                        & 128-bit \\
\hline
OKH                     &0.018  &0.033   &0.122   &0.114  &0.258    &0.011  &0.025   &0.104   &0.094
                        &0.091    \\
\hline
SketchHash              &0.052  &0.120   &0.202   &  -    &  -      &0.045  &0.108   &0.186   &  -
                        &-   \\
\hline
AdaptHash               &0.028  &0.097   &0.195   &0.222  &0.229    &0.029  &0.089   &0.178   &0.243
                        &0.285      \\
\hline
OSH                     &0.018  &0.021   &0.022   &0.043  &0.164    &0.010  &0.012   &0.014   &0.029
                        &0.132     \\
\hline
MIHash                  &\textbf{0.094}  &\textbf{0.191}   &0.244   &0.308  &0.332
                        &\textbf{0.083}  &\textbf{0.179}   &0.245    &0.282   &0.313   \\
\hline
\hline
HCOH                    &0.049  &0.173   &\textbf{0.259}    &\textbf{0.321}  &\textbf{0.347}
                        &0.063  &0.147   &\textbf{0.256}    &\textbf{0.298}  &\textbf{0.324}   \\
\hline
\end{tabular}
\end{table*}
\begin{figure*}[!t]
\begin{center}
\includegraphics[height=0.205\linewidth]{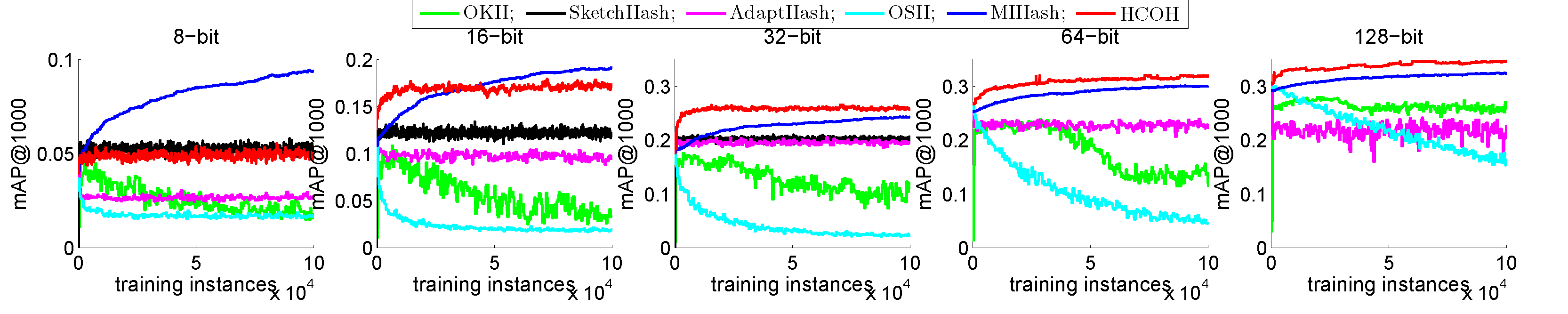}
\vspace{-2em}
\caption{\label{places_map} \textit{m}AP performance with respect to different number of training instances on Places205.}
\end{center}
\end{figure*}
\begin{figure}[!t]
\begin{center}
\includegraphics[height=0.63\linewidth]{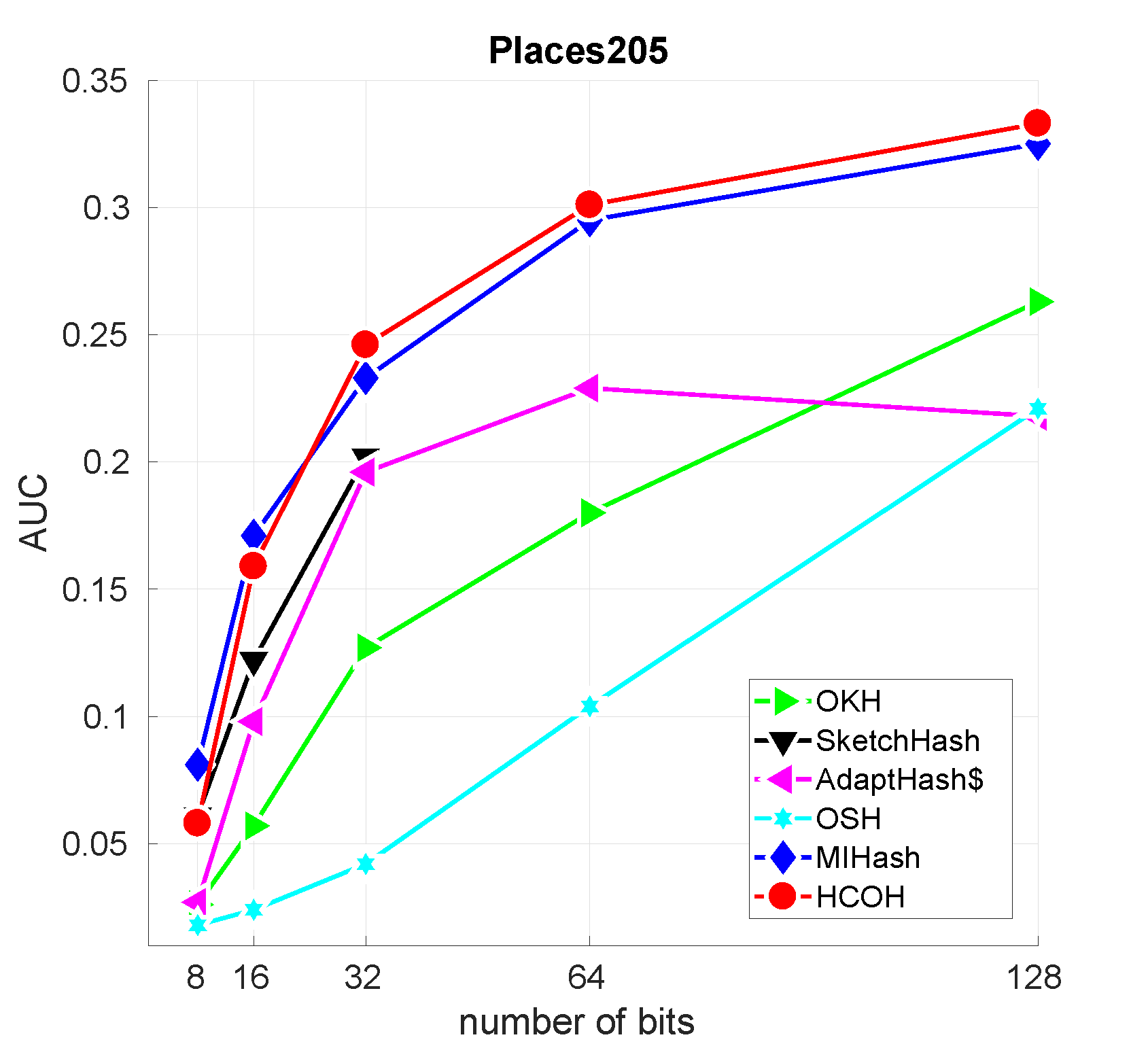}
\caption{\label{auc_places} AUC performance with respect to different lengths of hash bits on Places205.}
\vspace{-1.5em}
\end{center}
\end{figure}

\subsection{Parametric Settings}
We describe the parameters to be tuned during the experiments.
Due to that we share the common dataset configurations on CIFAR-10 and Places205, we directly adopt the parameters as described in \cite{fatih2017mihash} for all baselines.
Otherwise, we partition a validate set from training set with a size of 2K, 10K, 2K for CIFAR-10, Places205 and MINST, respectively, and identify the best choice for each parameter.
The following describes our parameter settings.
\begin{itemize}
\item \textbf{OKH}: The tuple $(C, \alpha)$ is set as $(0.001, 0.3)$, $(0.0001, 0.7)$ and $(0.001,0.3)$ for CIFAR-10, Places205 and MNIST, respectively.
\item \textbf{SketchHash}: The tuple $(sketch size, batch size)$ is set to \redliu{$(200,\allowbreak 50)$}, $(100, 50)$ and $(200, 50)$ for CIFAR-10, Places205 and MNIST, respectively.
\item \textbf{AdaptHash}: The tuple $(\alpha, \lambda, \eta)$ is set as $(0.9,0.01,0.1)$, \redliu{$(0.9,\allowbreak 0.01,0.1)$} and $(0.8, 0.01, 0.2)$ for CIFAR-10, Places205 and MNIST, respectively.
\item \textbf{OSH}: For all datasets, $\eta$ is set to $0.1$ and the ECOC codebook $C$ is populated the same way as in \cite{cakir2017online}.
\item \textbf{MIHash}: The tuple $(\theta, \mathcal{R}, A)$ as $(0, 1000, 10)$, $(0, 5000, 10)$ and $(0, 1000, 10)$ for CIFAR-10, Places205 and MNIST, respectively.
\end{itemize}

Due to the page limit, we do not explicitly explain the meaning of these parameters.
Detailed information can be found in the corresponding papers \cite{huang2013online, leng2015online, cakir2015adaptive, cakir2017online, fatih2017mihash}.
Also, for SketchHash, the batch size has to be larger than the length of hash bits.
Therefore, we do not report its performance when the hash bits are $64$ and $128$.

\subsection{Results and Discussions \label{sec4.4}}

\subsubsection{Results on CIFAR-10}
We first report the performance of the proposed  method on CIFAR-10.
Tab. \ref{cifar-map-precision500} illustrates the \textit{m}AP and Precision@500 of our method and the baselines with different hash bits.
\redliu{Fig. \ref{cifar-map} reports the \textit{m}AP with respect to different numbers of training instances, and Fig. \ref{auc_cifar} displays the corresponding AUC results when the hash bits are $8$, $16$, $32$, $64$ and $128$.}
%
We can see that the proposed method outperforms all of the other methods in all cases.

In terms of \textit{m}AP and Precision@500, we can observe that the proposed method achieves substantially better performance at all code lengths.
Comparing to the state-of-the-art method, \textit{i.e.}, MIHash, the proposed method shows a relative increase of $4.69\%$, $9.06\%$, $1.93\%$, $8.55\%$, $10.54\%$ for \textit{m}AP when hash bits are $8$, $16$, $32$, $64$, $128$, respectively, as well as $13.57\%$, $6.97\%$, $1.61\%$, $4.47\%$, $4.56\%$ for Precision@500 when hash bits are $8$, $16$, $32$, $64$, $128$, respectively.
As for the \textit{m}AP performance with respect to different numbers of training instances, the proposed method not only surpasses the state-of-the-art methods by a large margin, but also achieves satisfactory performance with less training instances. Taking the metric under 64-bit as an example, the proposed method obtains 0.6 \textit{m}AP when the training instances grow to around $1K$, while it takes MIHash nearly $8K$ instances to achieve the same \textit{m}AP.
For a deeper look, we further analyze the area under the \textit{m}AP curves in Fig. \ref{auc_cifar}.
Among all baselines including OSH that uses ECOC as a codebook, the proposed HCOH always achieves the best results, which implies that using Hadamard codebook \redliu{with our proposed online learning scheme} is more preferable than using ECOC with classical SGD learning scheme.

\begin{table*}[]
\centering
\caption{\textit{m}AP and Precision@500 Comparisons on MNIST with 8, 16, 32, 64 and 128 bits.}
\label{mnist-map-precision500}
\begin{tabular}{c|ccccc|ccccc}
\hline
\multirow{2}{*}{Method} & \multicolumn{5}{c|}{\textit{m}AP}                   & \multicolumn{5}{c}{Precision@500}          \\
\cline{2-11}
                        & 8-bit & 16-bit & 32-bit & 64-it & 128-bit & 8-bit & 16-bit & 32-bit & 64-bit
                        & 128-bit \\
\hline
OKH                     &0.100  &0.155   &0.224   &0.301  &0.404    &0.100  &0.257   &0.452   &0.606
                        &0.732    \\
\hline
SketchHash              &0.257  &0.312   &0.348   &  -    &  -      &0.420  &0.571   &0.641   & -
                        &-    \\
\hline
AdaptHash               & 0.138 &0.207   &0.319   &0.292  &0.208    &0.187  &0.368   &0.571   &0.558
                        &0.439    \\
\hline
OSH                     &0.130  &0.144   &0.130   &0.146  &0.143    &0.152  &0.171   &0.165   &0.222
                        &0.224    \\
\hline
MIHash                  &\textbf{0.664}  &\textbf{0.741}   &0.744   &0.713  &0.681
                        &\textbf{0.755}  &\textbf{0.801}   &0.823   &0.812  &0.812\\
\hline
\hline
HCOH                    &0.536  &0.708   &\textbf{0.756}   &\textbf{0.759}  &\textbf{0.771}
                        &0.662  &\textbf{0.801}   &\textbf{0.837}   &\textbf{0.848}  &\textbf{0.854}
\\
\hline
\end{tabular}
\end{table*}

\begin{figure*}[!t]
\begin{center}
\includegraphics[height=0.205\linewidth]{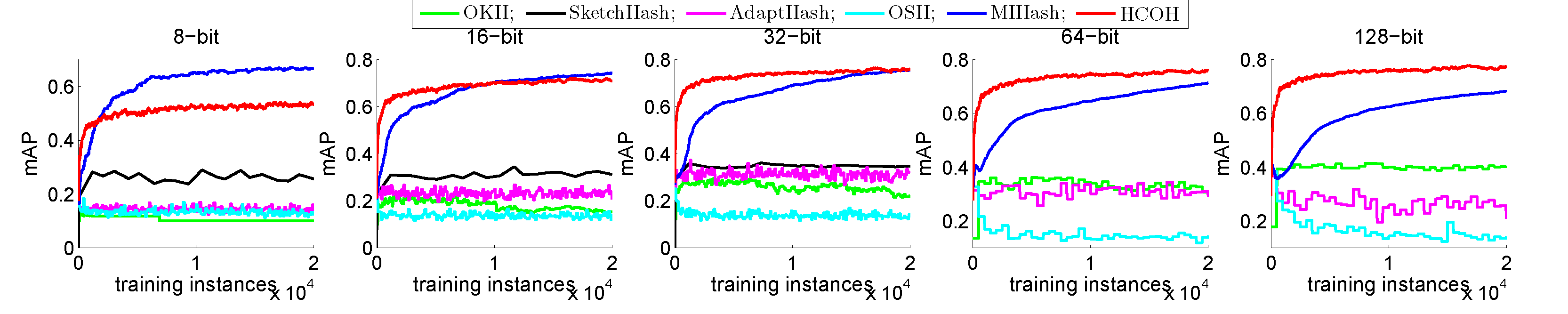}
\caption{\label{mnist_map} \textit{m}AP performance with respect to different number of training instances on MNIST.}
\end{center}
\end{figure*}

\begin{figure}[!t]
\begin{center}
\includegraphics[height=0.63\linewidth]{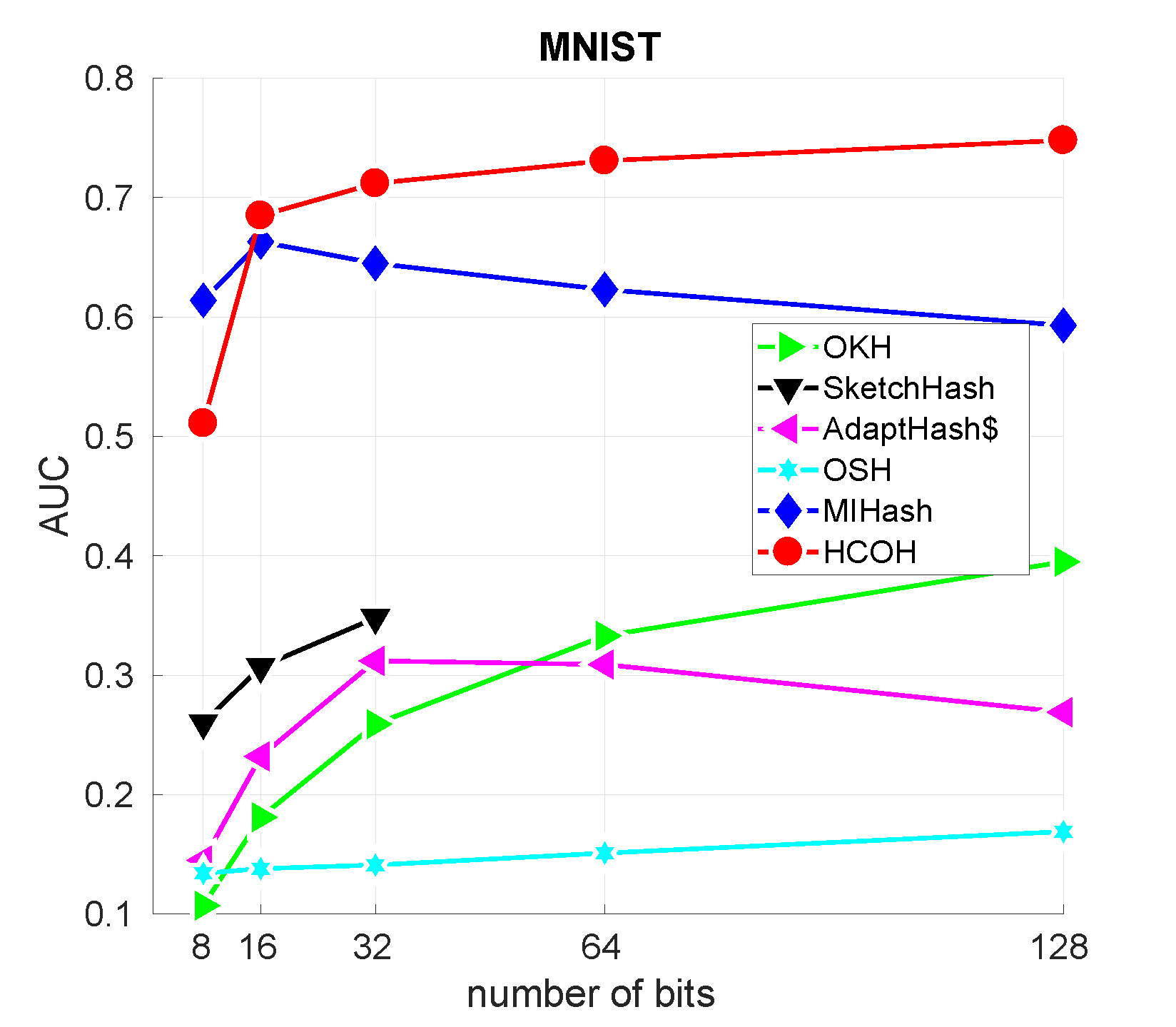}
\caption{\label{auc_mnist} AUC performance with respect to different lengths of hash bits on MNIST.}
\vspace{-1.5em}
\end{center}
\end{figure}

\subsubsection{Results on Places205}
Tab. \ref{places-map-precision500} shows comparative results about \textit{m}AP@1000 and Precision@500 on a larger-scale dataset Places205 with different code lengths, respectively.
The \textit{m}AP with respect to different numbers of training instances and its AUC are reported in Fig. \ref{places_map} and Fig. \ref{auc_places}.
We can find that in low bit cases, \textit{i.e.}, $16$-bit and $32$-bit, MIHash achieves the best results in all four metrics.
However, as the length of hash bit grows, the proposed HCOH still outperforms all the baselines including MIHash\redliu{, which claims that the overall performance of the HOCH is better on Places205}.

In detail, in low bits of $8$ and $16$, compared with the proposed HCOH, MIHash acquires $23.88\%$, $4.66\%$ gains for \textit{m}AP@1000 and $31.75\%$, $21.77\%$ gains for Precision@500, respectively.
Notably, under the setting of $8$-bit, SketchHash is second best.
When the hash bits are 32, 64 and 128, the proposed HCOH surpasses MIHash by $6.15\%$, $4.22\%$, $4.52\%$ gains for \textit{m}AP@1000 and $4.49\%$, $5.67\%$, $3.51\%$ gains for Precision@500, respectively.
We further look into the \textit{m}AP with respect to different number of training instances and the AUC under different hash bits.
As depicted in Fig. \ref{places_map}, when the lengths of hash codes are $8$ and $16$, at first, the proposed HCOH rapidly increase, and as the dataset grows, MIHash transcends.
However, when referring to $32$-bit, $64$-bit and $128$-bit, the proposed HCOH keeps the first throughout the training process.
In Fig. \ref{auc_places}, it shows similar results as in Fig. \ref{places_map}. In the $8$-bit, $16$-bit settings, MIHash performs the best.
Whereas, In the settings of $32$-bit, $64$-bit, $128$-bit, the proposed HCOH shows its dominance and keeps the first results.

To analyze the reason for unsatisfactory performance of the proposed HCOH in low hash bits, it owes to the usage of LSH to reduce the codeword's length, because LSH needs longer codes to achieve the theoretic convergence guarantee \cite{gionis1999similarity}.
Nevertheless, we argue that when facing large-scale datasets, longer binary codes are necessities to guarantee good performance.
Even though MIHash shows best in hash bits of $8$ and $16$, the performance are far from satisfying.
For example, in 8-bit setting, the $m$AP is only $0.094$, which is insufficient in real applications.
However, when hash bit is $64$, the best result increases to $0.321$, which is more applicable.
Therefore, longer binary codes are necessary to achieve workable performance in large-scale settings.

\subsubsection{Results on MNIST}
The \textit{m}AP and Precision@500 for all methods on MNIST are listed in Tab. \ref{mnist-map-precision500}.
The \textit{m}AP with regard to different numbers of training instances and AUC \redliu{curves} are displayed in Fig. \ref{mnist_map} and Fig. \ref{auc_mnist}\redliu{, respectively}.
We can observe that the results on MNIST are very similar to those on Places205.
When it comes to low hash bits of $8$ and $16$, the performance of the proposed HCOH is worse than MIHash.
But in most cases, our method surpasses all of the other methods.
As aforementioned, low hash bits are not suitable for large-scale datasets, especially for streaming data due to its low performance.

In particular, when the hash bits are $8$ and $16$, MIHash gets $23.88\%$ and  $0.05\%$ \textit{m}AP gains to the proposed method.
When hash bits are $32$, $64$ and $128$, the proposed HCOH gets $1.61\%$, $6.45\%$, $13.22\%$ \textit{m}AP gains to MIHash.
As for the Precision@500, MIHash gains $14.05\%$ higher than HCOH when the hash bit is $8$.
Both the proposed HCOH and MIHash earn $0.801$ \textit{m}AP under the hash bit of 16.
Regarding to the hash bits of $32$, $64$, $128$, the proposed method consistently outperforms MIHash by $1.70\%$, $4.43\%$ and $5.17\%$ \textit{m}AP gains, respectively.
Further, we analyze the \textit{m}AP with respect to different numbers of training instances and the corresponding AUC. %
As shown in Fig. \ref{mnist_map}, when the hash bit is $8$, the proposed method shows best \textit{m}AP at first, but is surpassed by MIHash as the number of instances increases.
Fig. \ref{auc_mnist} reports the same observation for AUC.
Interestingly, in regard of the 16-bit case, even MIHash transcends the proposed HCOH in the end, but the proposed HCOH still holds the first position for AUC metric.
For higher hash bits, the proposed method significantly surpasses all baselines by a large margin.

Based on Tab. \ref{cifar-map-precision500}, Tab. \ref{places-map-precision500} and Tab. \ref{mnist-map-precision500}, it can be observed that when the hash bits are $8$ and $16$, HCOH is worse than MIHash on Places205 and MNIST, while HCOH performs better on CIFAR-10.
We argue that this is owing to the dimensionality of features.
As introduced in Sec \ref{datasets}, features in CIFAR-10 are 4096-D, while it is only 128-D and 784-D for Places205 and MNIST, respectively.
So as to preserve mutual information in Hamming space, MIHash learns binary codes via linear mappings which suffers great quantization \dele{loss}\redliu{error} when mapping data from high-dimensional space into low-bit Hamming space.
Hence, we argue that when learning low-bit binary codings, $\mathbf{i.e.}$, $8$ or $16$ bits, our method is suitable for high-dimensional features, while MIHash is proper to low-dimensional features.

\subsubsection{Parameter Sensitivity Analysis}
We further analyze the \dele{potential} influence of hyper-parameters on the performance.
Unlike other baselines, there are only two parameters in the proposed HCOH, \textit{i.e.}, the learning rate $\eta$ and the batch size $n$, which reflects another advantage of the proposed HCOH\redliu{,} because less parameters denote simpleness to complement and less time spent on deciding optimal values.
For simplicity, at each round $t$, we set $\eta^t$ and $n^t$ as constants.

\begin{figure}[t]
\begin{center}
\begin{minipage}[t]{0.25\linewidth}
\centerline{
\subfigure[\label{lr_map} \textit{m}AP with varying learning rates.]{
\includegraphics[width=1.8\linewidth]{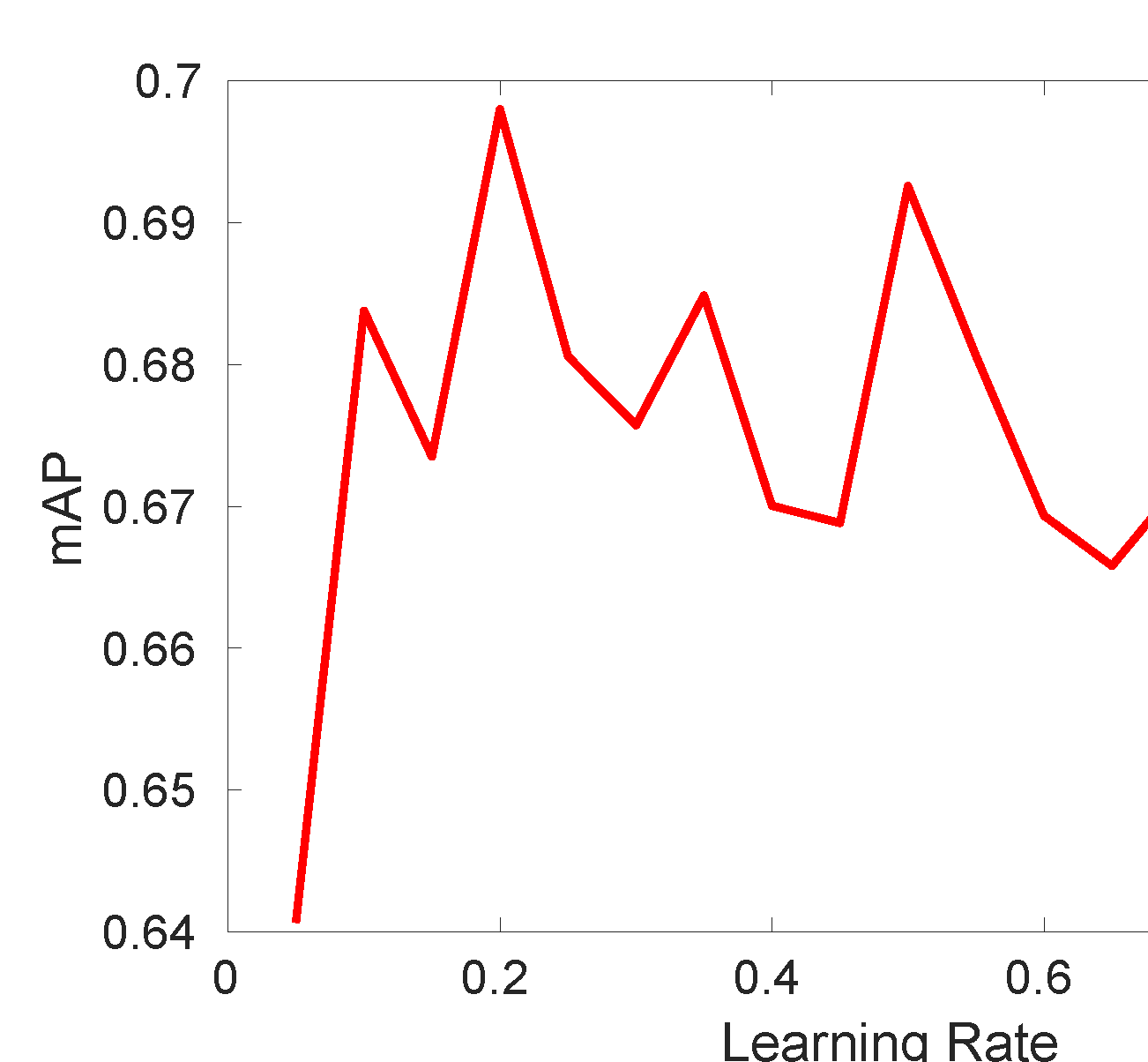}}\hspace*{-0.05\linewidth}
\subfigure[\label{batch_map} \textit{m}AP over different sizes of batch.]{
\includegraphics[width=1.8\linewidth]{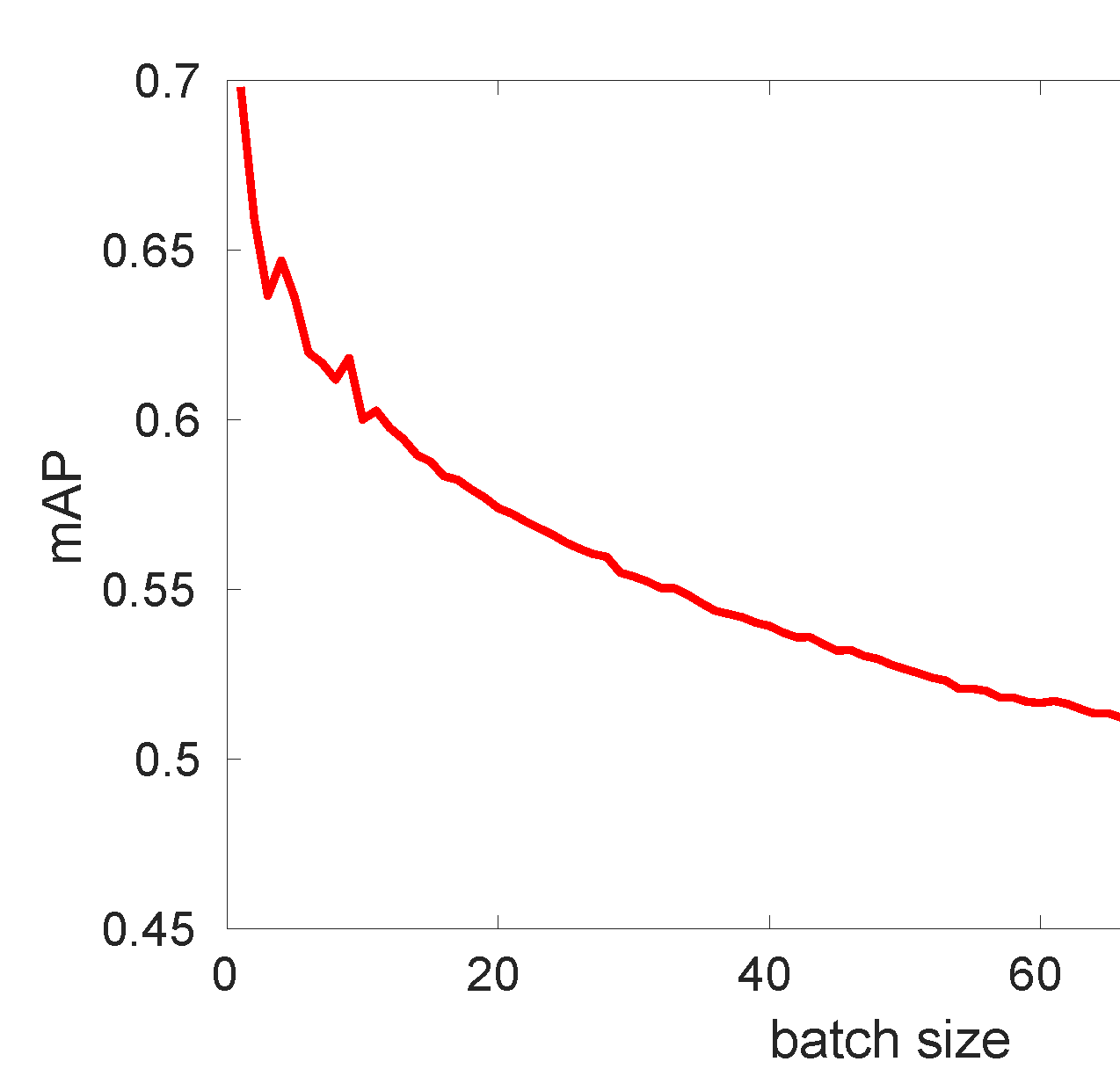}}
}
\end{minipage}
\end{center}
\vspace{-1em}
\caption{ The analysis of hyper-parameters.}
\vspace{-2.3em}
\end{figure}

To validate the effectiveness of these two parameters, we conduct experiments on CIFAR-10 when the hash bit is 16.
In Fig. \ref{lr_map}, we plot the \textit{m}AP curves under different values of learning rate.
We can see that the \textit{m}AP results fluctuate as the learning rate $\eta$ varies. This is owing to the random sampling process involved in the evaluation protocol.
In general, when $\eta = 0.2$, the proposed HCOH performs the best. Fig. \ref{batch_map} shows the \textit{m}AP performance along with the increase of batch size $n$.
Generally, the performance of the proposed HCOH degenerates with the increase of $n$.
The precise results for $n^t=1$ and $n^t=2$ are 0.755 and 0.716, respectively.
This is because individual update preserves more instance-level information.
The best choice for $n$ is $1$ in such a case.

\dele{Similarily}\redliu{Similarly}, the same experiments can be conducted for Places205 and MNIST.
In this paper, the tuple ($\eta$, $n$) is set as ($0.2$,$1$), ($0.1$,$1$) and ($0.2$, $1$) for CIFAR-10, Places205 and MNIST, respectively.
Through \redliu{the }\dele{this} analysis, we demonstrate that the proposed HCOH only needs one instance to update the hash functions each round, which differs HCOH to most OH methods, which needs at least two instances.


%
\subsection{Time Complexity}
To demonstrate the efficiency of the proposed HCOH, we further compare the training time on CIFAR-10 when the hash bit is $32$.
The results are summarized in Tab. \ref{training-time}\dele{. As shown in Tab. \ref{training-time},}\redliu{ , which shows that} HCOH has acceptable training time.
Although OKH and SketchHash are more efficient, they suffer from unsatisfactory performance as analyzed in Sec. \ref{sec4.4}.
When comparing to OSH, the proposed HCOH gains a $20.42\times$ training speed acceleration.
To analyze, though OSH also adopts codebook-based scheme, it has to utilize the boosting algorithm to improve the performance, which increases the training time.
Regarding to MIHash, the proposed HCOH obtains a $5.26\times$ training speed acceleration, which is due to the usage of mutual information, \textit{i.e.}, given a query, MHIash has to calculate the Hamming distance between its neighbors and non-neighbors.
Even if in some situations of low hash bits, the performance of MIhash may surpass the proposed HCOH, it has unavoidably introduced more training time, which is in many cases unacceptable in \redliu{OH}\dele{online learning} .

\begin{table}[]
\centering
\caption{Training times on the CIFAR-10 with 20K training instances, under 32-bit hash codes.}
\label{training-time}
\begin{tabular}{cc}
\hline
Method     & Training Time (s) \\
\hline
\hline
OKH        & 12.40             \\
\hline
SketchHash & 8.25              \\
\hline
AdaptHash  & 78.52             \\
\hline
OSH        & 333.70            \\
\hline
MIHash     & 95.24             \\
\hline
Ours       & 15.58             \\
\hline
\end{tabular}
\end{table}

\section{CONCLUSION} \label{sec5}
In this paper, we propose a robust supervised online hashing scheme, termed Hadamard Codebook based Online Hashing \dele{(HCOH)}, which can be trained very efficiently online, and is not limited by predefining the category number of the streaming dataset.
To this end, the proposed HCOH is firstly associated with a codebook sampled from the generated Hadamard matrices, and then designates the codeword in the codebook as the centroid of data sharing the same label space, so as to conduct the learning of hash functions.
To keep consistency with the length of hash bits, locality sensitive hashing is further employed to reduce the codeword dimension. Stochastic gradient descend \dele{(SGD)} is developed to update the hash codes for streaming data online.
In optimization, the proposed HCOH only needs one training instance each round.
Extensive experiments with quantitative evaluation metrics and benchmarks including CIFAR-10, Places-205, and MNIST demonstrate the merits of of the proposed method over the state-of-the-art.

\section{Acknowledge}
This work is supported by the National Key R\&D Program (No. 2017YFC0113000, and No. 2016YFB1001503), Nature Science Foundation of China (No. U1705262, No. 61772443, and No. 61572410), Post Doctoral Innovative Talent Support Program under Grant BX201600094, China Post-Doctoral Science Foundation under Grant 2017M612134, Scientific Research Project of National Language Committee of China (Grant No. YB135-49), and Nature Science Foundation of Fujian Province, China (No. 2017J01125 and No. 2018J01106).

\bibliographystyle{ACM-Reference-Format}
\balance
\bibliography{mybib}

\end{document}